\documentclass[conference]{IEEEtran}
\IEEEoverridecommandlockouts

\usepackage{cite}
\usepackage{amsmath,amssymb,amsfonts}
\usepackage{algorithmic}
\usepackage{graphicx}
\usepackage{textcomp}
\usepackage{xcolor}

\usepackage{makecell}
\usepackage{booktabs}
\usepackage{tabularx}
\usepackage{multirow}
\newcolumntype{L}[1]{>{\hsize=#1\hsize\raggedright\arraybackslash}X}
\newcolumntype{R}[1]{>{\hsize=#1\hsize\raggedleft\arraybackslash}X}
\newcolumntype{C}[1]{>{\hsize=#1\hsize\centering\arraybackslash}X}

\usepackage[bookmarks=False,
colorlinks,
urlcolor=blue,
citecolor=black!50!green,
linkcolor=blue
]{hyperref}

\usepackage{cleveref}

\usepackage{tikz}
\usetikzlibrary{calc, shapes, fit}
\usepackage{pgfplots}
\usepackage{pgfplotstable}

\begin{document}

\title{Recognition-free Question Answering on Handwritten Document Collections}

\author{\IEEEauthorblockN{Oliver T{\"u}selmann, Friedrich M{\"u}ller, Fabian Wolf and Gernot A. Fink}
	\IEEEauthorblockA{\textit{Department of Computer Science} \\
		\textit{TU Dortmund University}\\
		44221 Dortmund, Germany \\
		{firstname.lastname}@cs.tu-dortmund.de}
}

\maketitle

\begin{abstract}
In recent years, considerable progress has been made in the research area of Question Answering (QA) on document images. Current QA approaches from the Document Image Analysis community are mainly focusing on machine-printed documents and perform rather limited on handwriting.
This is mainly due to the reduced recognition performance on handwritten documents.
To tackle this problem, we propose a recognition-free QA approach, especially designed for handwritten document image collections. 
We present a robust document retrieval method, as well as two QA models.
Our approaches outperform the state-of-the-art recognition-free models on the challenging BenthamQA and HW-SQuAD datasets. 
\newline

\end{abstract}


\section{Introduction}
Question Answering (QA) is still an open and major research topic in a wide variety of disciplines \cite{Wu17,Zhu21,Mathew21D}. 
Especially, the communities of Computer Vision (CV) and Natural Language Processing (NLP) focus on this task and made considerable progress \cite{Wu17,Zhu21}. 
Over the last few years, the Document Image Analysis (DA) community has shown an increasing interest in QA \cite{Mathew21,Mathew21D}.  
The majority of DA approaches tackle this task by adapting and using models from the NLP and CV communities \cite{Xu20,Mathew21,Powalski21}. 
Thereby, the text from a document image is transcribed and an answer is determined using a textual QA system  \cite{Xu20,Mathew21}.
This already leads to high performances for machine-printed document images with low recognition error rates \cite{Mathew21D}. 
However, the performances of these approaches decrease considerably on handwritten document images \cite{Mathew21,Mathew21D}.
This is mainly due to the considerably reduced recognition accuracy, 
even though, substantial progress has been made in handwritten text recognition (HTR) over the last few years \cite{Kang19}. 

\begin{figure*}[t]
	\centering
	\includegraphics[width=1\linewidth]{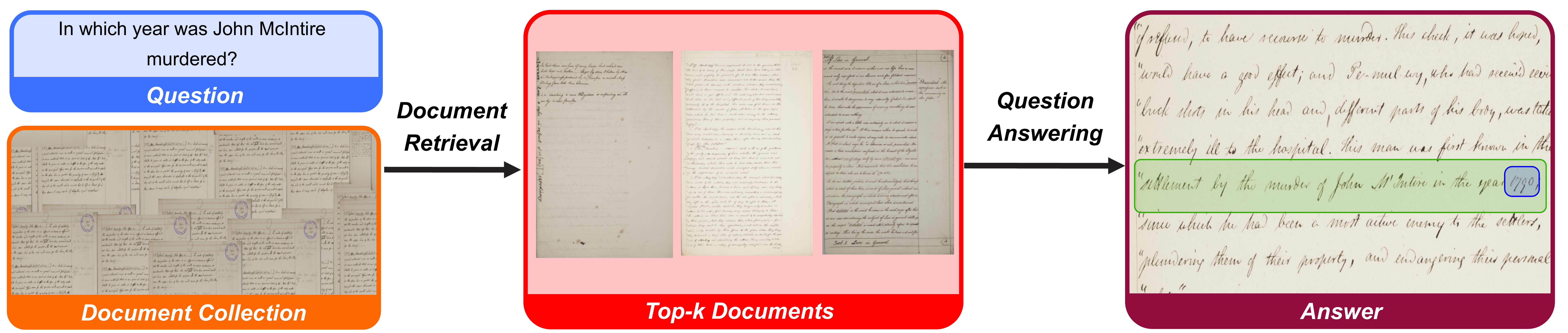}
	\caption{An overview of the Question Answering pipeline on document image collections. Given a textual question and a document image collection, a document retriever identifies the $k$ most relevant documents from the collection for answering the question. Finally, a word (blue) or line (green) region from one of these $k$ document images is returned as the answer. }
	\label{fig:overview}
\end{figure*}

\vspace{0.2cm}
Answering questions on handwritten document images requires models that are robust with respect to handwriting recognition errors or do not rely on textual input. 
This is particularly important for QA on unknown document collections, as training data is usually not available and therefore, high handwriting recognition error rates are expected.
For developing and evaluating such approaches, Mathew et al. recently proposed the BenthamQA and HW-SQuAD datasets \cite{Mathew21}.
These datasets provide questions as strings in natural language and expect answers as image regions of a document image rather than a textual response. Finding an answer in a single document image is already a challenging task. However, it is even more complicated in real world scenarios as a large collection of document images is often given. Therefore, it is crucial to identify documents in the collection that are relevant to answer the question and afterwards extract the answer from these documents. Fig. \ref{fig:overview} provides an overview of this pipeline.

\vspace{0.2cm}
In this work, we propose a recognition-free approach for answering questions on handwritten document image collections. 
We present a robust document retriever as well as two QA approaches. The first QA model is based on the approach of Mathew et al. \cite{Mathew21} and replaces their aggregation strategy with an attention-based method. The second model is based on a QA architecture from the NLP domain and enables recognition-free QA on both word and line level. We compare our approach with recognition-free as well as recognition-based QA approaches on the challenging BenthamQA \cite{Mathew21} and HW-SQuAD \cite{Mathew21} datasets and are able to outperform state-of-the-art results by a large margin.

\vspace{0.2cm}
\section{Related Work}
QA on document collections usually requires a two-stage approach consisting of a document retriever and a QA model. We provide an overview on textual document retrieval (see Sec. \ref{ssec:retrieval}) and QA in the visual, textual as well as document image domain (see Sec. \ref{ssec:qa}).

\vspace{.2cm}
\subsection{Document Retrieval}
\label{ssec:retrieval}
Document retrieval is an information retrieval task that receives a textual request and returns a set of documents from a given document collection that best matches the query. Traditional approaches rely on counting statistics between query and document words \cite{Mitra18,Guo20}. Different weighting and normalization schemes over these counts lead to Term Frequency-Inverse Document Frequency (TF-IDF) models, which are still popular \cite{Mitra18,Guo20}. However, these models ignore the position of occurrences and the relationships with other terms in the document \cite{Guo20}. Therefore, models have been developed that can learn the relevance between questions and documents. Learning-To-Rank (LTR) \cite{Guo20} is a well-known document retrieval approach, which represents a query-document pair as a vector of hand-crafted features and trains a model to obtain similarity scores. Recently, deep neural ranking models outperform LTR models \cite{Guo20}. For a detailed overview on document retrieval, see \cite{Mitra18,Guo20}.

\vspace{.2cm}
\subsection{Question Answering}
\label{ssec:qa}
QA is applied in various domains, leading to large variations among approaches. We present an overview on purely textual QA approaches as well as Visual Question Answering (VQA) models from the CV domain. Furthermore, we discuss recent progress for VQA on document images.

\vspace{.3cm}
\subsubsection{Textual Question Answering}

The textual QA community is mainly focusing on the Machine Reading Comprehension (MRC) \cite{Zeng20} and OpenQA \cite{Zhu21} tasks. In MRC, only one document is given and the answer is a snippet of the document. There is also an extension for this task, whereby models have to decide whether a question is answerable based on the document \cite{Zeng20}. Traditional MRC approaches are mainly implemented based on handcrafted rules or statistical methods \cite{Zeng20}. Long Short-Term Memory (LSTM)-based models with Attention \cite{Seo17} achieved further progress in this field. In recent years, Transformer models (e.g. BERT \cite{Devlin19}) improved the results considerably \cite{Seo17}.
These models benefit from largely pre-trained word embeddings, which encode useful semantic information between words.
Currently, specialized transformer models (e.g. LUKE \cite{Yamada20}) lead to state-of-the-art results.  
In contrast to MRC, OpenQA tries to answer a given question without any specified context. It usually requires the system to search for relevant documents in a large document collection and generate an answer based on the retrieved documents. OpenQA models are mainly a combination of document retrieval and MRC-based approaches \cite{Zhu21}. For a detailed overview of textual QA, see \cite{Zhu21}.

\vspace{.3cm}
\subsubsection{Visual Question Answering}
Given an image and a query in natural language, Visual Question Answering (VQA) tries to answer the question using visual elements of the image and textual information from the query \cite{Wu17}. Most approaches rely on an encoder–decoder architecture, which embed questions and images in a common feature space \cite{Wu17}. This allows learning interactions and performing inference over the question and the image contents. Practically, image representations are obtained with Convolutional Neural Networks (CNNs) pre-trained on object recognition \cite{Wu17}. Text representations are obtained with word embeddings pre-trained on large text corpora \cite{Wu17}. RNNs are used to handle the variable size of questions \cite{Wu17}. Further progress in this field has been made using Attention \cite{Wu17}. The attention mechanism allows the model to assign importance to features from specific regions of the image. Recently, Transformer based architectures achieved state-of-the-art results on multiple VQA benchmarking datasets \cite{Wu17}. For a detailed overview of VQA, see \cite{Wu17}.

\vspace{.3cm}
\subsubsection{Document Image Visual Question Answering}
Mainly due to several new competitions\cite{Mathew21D,Mathew21E} and datasets \cite{Mathew21,Mathew21D,Mathew21E}, there has been major progress in the area of answering questions on document images. These datasets provide MRC \cite{Mathew21D,Mathew21E} as well as OpenQA tasks \cite{Mathew21}. The approaches and datasets mainly focus on machine-printed documents, which contain visual and structural information (e.g. charts, diagrams) \cite{Xu20,Powalski21,Mathew21}. The layout is important for answering most of the questions \cite{Xu20,Powalski21,Mathew21D}. The approaches are based on textual recognition results and adapt state-of-the-art QA systems from the NLP domain \cite{Xu20,Powalski21}. Recently, Mathew et al. \cite{Mathew21} published a first dataset for QA on handwritten document collections. Furthermore, they proposed a recognition-free QA approach, which outperforms recognition-based QA models on handwritten datasets \cite{Mathew21}.

\section{Method}
In this section, we present our recognition-free approach for answering questions on document image collections. The approach consists of a document retriever (see Sec. \ref{ssec:m_retrieval}) and a QA model (see Sec. \ref{ssec:m_qa}). Both models are based on the robust Pyramidal Histogram of Characters (PHOC) attribute representation (see Sec. \ref{ssec:m_phoc}). Given a query and a collection of word and line-segmented document images, our document retrieval approach assigns a score to each document image, indicating its relevance to answer the query. For each of the K most relevant documents, our QA model determines answer snippets and an associated certainty value. Finally, the snippet with the highest certainty value is returned as the answer.

\subsection{Query and document representation}
\label{ssec:m_phoc}
To compute a similarity score between a document image and a query, as well as for question answering, the question words and the document images have to be transformed into a vector representation. Since we follow a recognition-free approach and the question is provided in a textual and the documents in a visual form, we use the PHOC representation that allows a robust mapping of words and images into the same space. PHOC is an attribute representation that is successfully and widely used in the word spotting domain. We use the TPP-PHOCNet \cite{Sudholt16} to realize a mapping from word images to a PHOC representation. The representations of the word images are finally stored in the order of their occurrences in the document image.

\subsection{Retrieval}
\label{ssec:m_retrieval}
\begin{figure}
	\centering
	\includegraphics[width=1\linewidth]{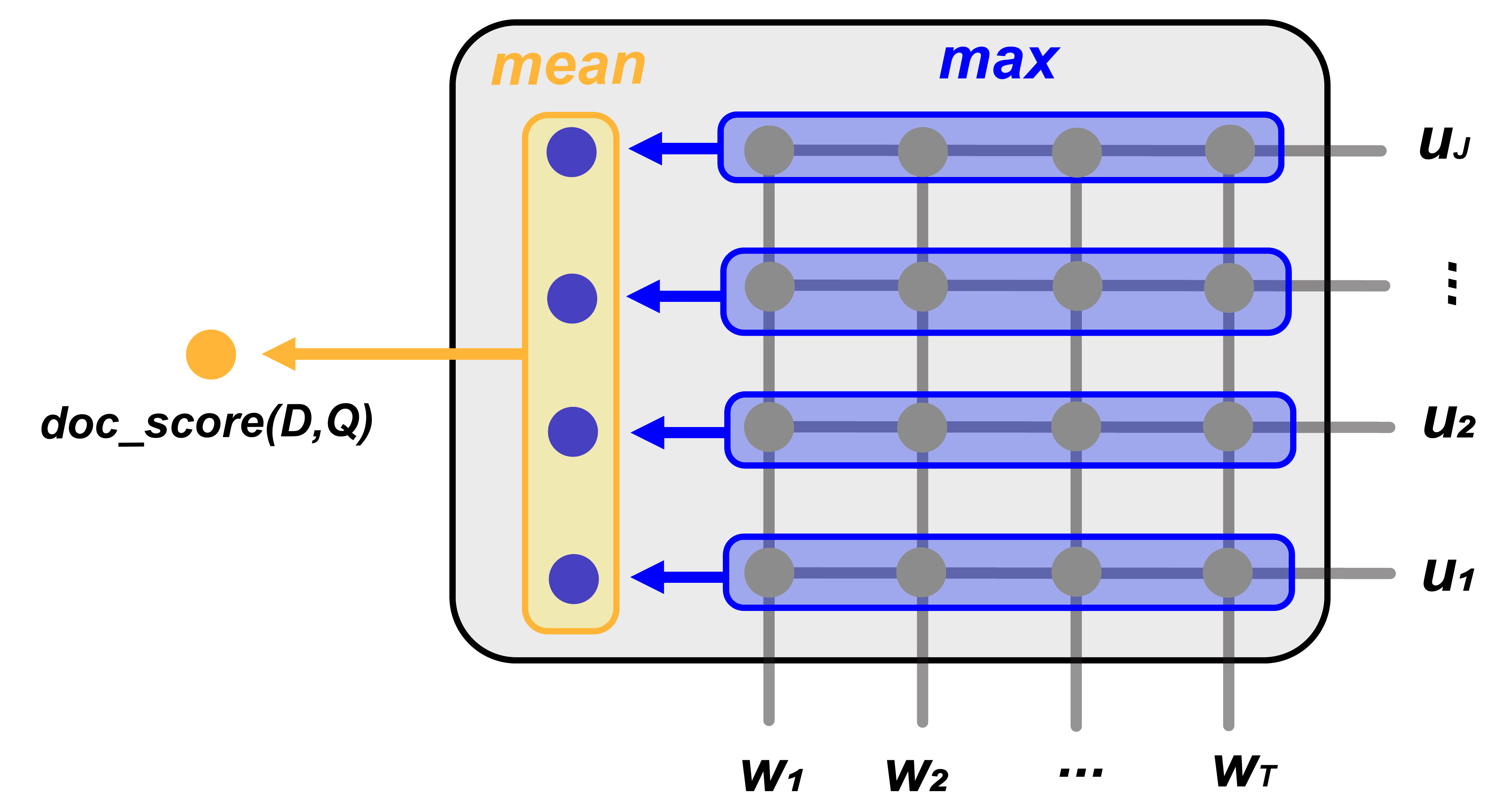}
	\caption{Our attention based retrieval approach for calculating the similarity between a query $Q=[u_1,...,u_J]$ and a document or snippet $D=[w_1,...,w_T]$.} 
	\label{fig:attention}
\end{figure}

To determine the most relevant documents regarding a query in a given collection, we follow a similar approach as described in \cite{Mathew21}. However, we do not aggregate all word images into one vector per document and then compare this to an aggregated query vector. Instead, we use the similarity between each word image from a document $D$ and each question word from the pre-processed query $Q$ as described in equation \ref{eq:docscore} and visualized in Fig. \ref{fig:attention}. We use the cosine similarity as the similarity measure. For each PHOC encoded question word $q \in Q$, the maximum similarity between $q$ and the predicted PHOC vectors of the word images $w \in D$ is calculated. The overall similarity between $Q$ and $D$ is the averaged value over all these similarity scores and is computed for each document in the collection. Finally, the documents from the collection are sorted in descending order with respect to the calculated scores and the first K documents are returned as the result. In the following, we denote this approach as \textit{Attention-Retriever}.

\begin{equation}\label{eq:docscore}
	doc\_score(D,Q) = \frac{1}{|Q|} * \sum_{q \in Q}\max_{w \in D} [sim(w,q)]
\end{equation}

\begin{figure*}
	\centering
	\includegraphics[width=\linewidth]{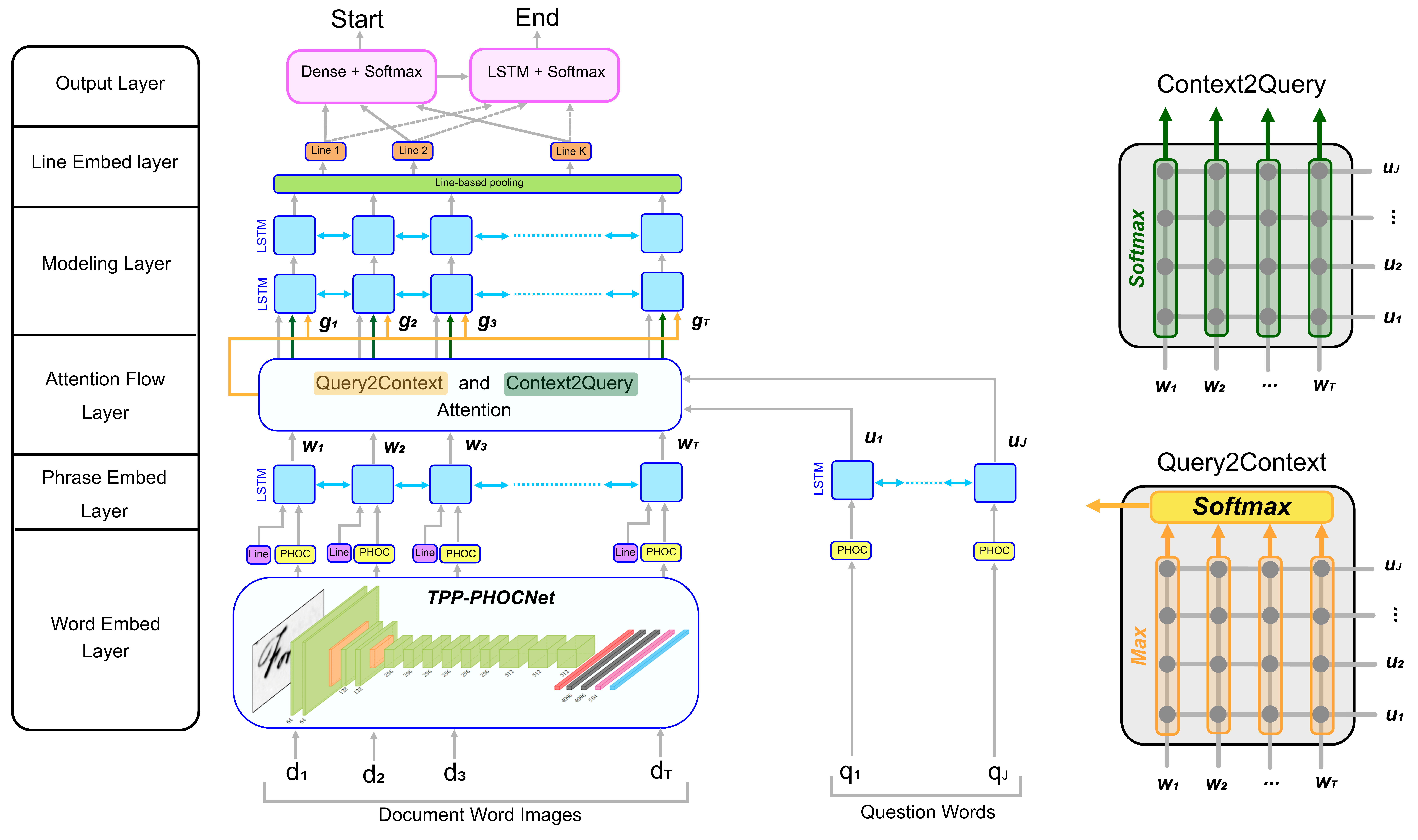}
	\caption{The adapted BIDAF architecture for recognition-free Question Answering on line level.}
	\label{fig:bidaf}
\end{figure*}

\subsection{Question Answering}
\label{ssec:m_qa}

The recognition-free QA approach from \cite{Mathew21} transforms document images into a set of two-line image snippets. For each of these snippets, an aggregated vector representation is determined based on the corresponding word images and is used to compute a similarity score with respect to an aggregated query vector. The score represents the confidence of finding the answer in the corresponding document region and determines the final answer of the system. Even though this approach can correctly locate the answers for some questions, the intuition behind this method is fairly questionable. The approach does not learn any real relationship between context and question, but exploits the heuristic that question words often occur close to the answer. Therefore, the approach does not realize a classical QA system, but rather an adapted syntactic word spotting approach for snippets.

NLP models are mainly based on the successes in transfer learning, where contextualized word embedding models were pre-trained on very large text collections.
Unfortunately, transfer learning on handwritten word images is currently difficult and a robust mapping of word images into a semantic space is challenging even for static semantic word embeddings \cite{Tueselmann20}. Therefore, it is currently not straightforward to adapted state-of-the-art NLP approaches to this task. However, there are previous state-of-the-art QA models from the NLP domain that do not rely on contextualized word embeddings and still lead to high performances on most datasets.

We follow the approach of the textual Bidirectional Attention Flow for Machine Comprehension (BIDAF) \cite{Seo17} model from the NLP domain and adapt it to a recognition-free QA model working on line instead of word level (see Fig. \ref{fig:bidaf}). All word images from a given document as well as the textual question words are represented as PHOC vectors. We further encode the line correspondence of each word image in the document using the positional encoding strategy from \cite{Devlin19} and append them to the corresponding PHOC representations.
BLSTM extracts and models the relations between the word image representations from a given document as well as for the question word representations. Afterwards, attention scores between the obtained context $(w_1,...,w_T)$ and question $(u_1,...,u_J)$ vectors are determined. The word representations are concatenated with their attention weighted versions and serve as the input to another two-stage BLSTM architecture, which models the relationship between questions and contexts. In this process, the BLSTM has as many outputs as the number of words in the document. The outputs of the BLSTM are reduced to the number of lines in the document by summing the word representations according to their line membership in the document. A dense layer is applied to each of these line representations and a softmax operation is performed. The result represents the pseudo-probability distribution for the start line of the answer. For calculating a similar distribution for the end line, the line representations are fed into another BLSTM and a dense Layer as well as a softmax operation is applied to its output.
The confidence for the prediction is the sum of the values before the softmax operation for the predicted start and end line indices.

The architecture can be used for word-level predictions by removing the line embedding layer. In the following, we will refer to the line-level model as \textit{BIDAF-Line} and the word-level model as \textit{BIDAF-Word}. In addition, we will refer to the adapted recognition-free QA approach of \cite{Mathew21} as \textit{Attention-QA}.

\section{Experiments}
We evaluate our proposed recognition-free QA approach on the HW-SQuAD and BenthamQA datasets (see Sec. \ref{ssec:dataset}). Sec. \ref{ssec:details} presents the implementation details and Sec. \ref{ssec:eval} discusses the evaluation results. The performance of QA systems is measured using the Double Inclusion Score (DIS) as introduced in \cite{Mathew21} and shown in equation \ref{eq:dis}. The Small Box (SB) includes the word images that contain the answer. The Large Box (LB) includes all word images from those lines that are part of the SB as well as those from the lines above and below it. The Answer Box (AB) includes the word images from the lines predicted by the QA system. The prediction is considered a correct answer if the score is above 0.8.

\begin{equation}\label{eq:dis}
	DIS = \frac{AB \cap SB}{|SB|}\times\frac{AB \cap LB}{|AB|}
\end{equation}

\subsection{Dataset}
\label{ssec:dataset}
We train and evaluate our models on the recently proposed HW-SQuAD and BenthamQA datasets \cite{Mathew21}, which contain question-answer pairs on handwritten document image collections in the English language. The datasets vary considerably in their size and characteristics and include synthetically generated as well as real handwritten documents.

BenthamQA \cite{Mathew21} is a small historical handwritten QA dataset where questions and answers were created using crowdsourcing.
The historic dataset contains 338 documents written by the English philosopher Jeremy Bentham and shows some considerable variations in writing styles. 
The dataset provides only a test set consisting of 200 question-answer pairs on 94 document images. The remaining 244 documents from the collection are used as distractors.

HW-SQuAD \cite{Mathew21} is a QA dataset on syntactically generated handwritten document images from the textual SQuAD1.0 \cite{Rajpurkar16} dataset. The textual dataset is actually defined for an MRC task and was adapted by Mathew et al. \cite{Mathew21} to an OpenQA task. The synthetic dataset consists of 20963 document pages containing a total of 84942 questions. The official partitioning splits the dataset into 17007 documents for training, 1889 for validation and 2067 for testing. Thereby, the training, validation and test sets contain 67887, 7578 and 9477 questions respectively.

\subsection{Implementation Details}
\label{ssec:details}
Our proposed document retriever relies on pre-segmented word images and our QA approaches also need line annotations. For our experiments, we use the gold-standard word and line bounding boxes available with the datasets. Questions are split into words and stopwords are removed using NLTK \cite{Bird06}. For training the BIDAF architecture, we use the HW-SQuAD dataset. We do not change the proposed parameters from \cite{Seo17} and use a hidden layer size of 100 as well as dropout with probability 0.2 for the BLSTMs. For optimization, we use ADADELTA \cite{Zeiler12} with the Cross Entropy loss and a learning rate of 0.5. The positional line encoding produces a 30-dimensional vector using sine and cosine functions.

For word representation, we use a 504-dimensional PHOC vector consisting of lowercase letters (a-z), numbers (0-9) and the levels 2, 3, 4 and 5. We pre-train the TPP-PHOCNet on the HW-SQuAD \cite{Mathew21} as well as IIIT-HWS \cite{Krishnan19} datasets and fine-tune the model on the IAM database\cite{Marti02}. We use a batch size of 40 and a momentum of 0.9. The parameters of the network are updated using the Stochastic Gradient Decent optimizer and the Cosine loss. The learning rate is set to 0.01 during pre-training and 0.001 while fine-tuning. It is divided by two if the loss has not decreased in the last three epochs. We binarize the word images to remove the background from text. This is especially important as the background of document images from BenthamQA largely differ from those in IIIT-HWS and IAM.

\subsection{Results}
\label{ssec:eval}
In this section, we show the evaluation performances of our recognition-free QA approach on handwritten document image collections. We evaluate and compare the document retrieval approach in Sec. \ref{sssec:retrieval} and the three QA approaches in Sec. \ref{sssec:qa}. Finally, we present and discuss the results on the combination of the retrieval and QA approaches in Sec. \ref{sssec:e2e}. For all subsequent experiments, we use the Attention-Retriever approach presented in Sec. \ref{ssec:m_retrieval}. 

\vspace{.3cm}
\subsubsection{Retrieval}
\label{sssec:retrieval}
For answering a given question in a document collection, it is common to determine the five most relevant documents with respect to the query.
To evaluate those retrieval models, we use the Top-5 accuracy as described in \cite{Mathew21}. This score represents the percentage of questions from a given test set that have their associated answer document in the top five predicted retrieval results. In table \ref{tab:retrieval} we present the Top-5 accuracy scores for our document retrieval approach and compare it to the literature. We also show the results of annotation-based NLP models in this table.  

The results show that we are able to improve the state-of-the-art Top-5 accuracy scores on both datasets. We clearly outperform the recognition-free approach from the literature. On the HW-SQuAD dataset, we perform marginally better compared to the recognition-based approach proposed in \cite{Mathew21}. The recognition-based model performs on nearly perfect recognition results (97.9\% word accuracy). When the recognition performance becomes worse as in BenthamQA (23.2\% word accuracy), the vulnerability of the approach to recognition errors is revealed and only a low performance can be achieved. The results from our model with ground truth and predicted PHOCs are almost identical, demonstrating the robustness of our approach. The differences between HW-SQuAD and BenthamQA can be explained by the lower prediction performance of the PHOC vectors for BenthamQA. In this case, the TPP-PHOCNet can achieve a query-by-string score of 98.5 on HW-SQuAD and 77.3 on BenthamQA. Interestingly, the annotation-based NLP method can only achieve marginally higher scores compared to our attention approach, demonstrating the capabilities of our model. 

\begin{table}[t]
	\centering
	\caption{Top-5 accuracy (\%) for document retrieval approaches}
	\label{tab:retrieval}
	\begin{tabular}{clcc}
		\toprule
		&Approach & HW-SQuAD	& BenthamQA\\ 
		\midrule
		\parbox[t]{2mm}{\multirow{2}{*}{\rotatebox[origin=c]{90}{Pred.}}}& Mathew et al. (rec-free) \cite{Mathew21} & 46.5	& 55.5 \\
		& Mathew et al. (rec-based)  \cite{Mathew21} & 86.1 & 32.0	\\ 
		& Attention-Retriever & \textbf{86.2}	&  \textbf{92.5} \\	
		\midrule
		\midrule
		\parbox[t]{2mm}{\multirow{2}{*}{\rotatebox[origin=c]{90}{GT}}}& Mathew et al. (rec-based)  \cite{Mathew21} & 90.2	&  98.5 \\	
		& Attention-Retriever & 87.2	&  98.0 \\	
		\bottomrule
	\end{tabular}
\end{table}

\vspace{.3cm}
\subsubsection{Question Answering}
\label{sssec:qa}
\begin{table}[t]
	\centering
	\caption{Machine Reading Comprehension. Performance measured in DIS.}
	\label{tab:qa}
	\begin{tabular}{clcc}
		\toprule
		& QA-Approach & HW-SQuAD	& BenthamQA\\ 
		\midrule
		\parbox[t]{2mm}{\multirow{3}{*}{\rotatebox[origin=c]{90}{Pred.}}} & Attention-QA  & 47.5	&  38.5\\	
		& BIDAF-Word  & 57.2	&  28.0\\ 
		& BIDAF-Line  & \textbf{68.1}	&  \textbf{50.5}\\	  
		\midrule
		\midrule
		\parbox[t]{2mm}{\multirow{4}{*}{\rotatebox[origin=c]{90}{GT}}} &Attention-QA & 47.7 & 39.0 \\
		& BIDAF-Word & 57.7 & 47.0\\ 
		& BIDAF-Line & 68.7 & 62.0\\
		& BERT \cite{Devlin19} & 94.4 & 88.0\\
		\bottomrule
	\end{tabular}
\end{table}
In order to evaluate the performance of our three proposed QA approaches without the influence of the retrieval model, we evaluate their performances on a MRC  rather than the OpenQA task. Thus, the QA systems only work on the document that contains the answer to the question. Table \ref{tab:qa} shows the results of our QA approaches as well as upper bounds using an annotation-based state-of-the-art QA approach (BERT \cite{Devlin19}).

The results show that our line-based BIDAF model can achieve higher scores compared to the word-based model and the attention based approach. A comparison with the literature is not possible, as Mathew et al. do not evaluate their approaches on this task. As already shown for the document retrieval, a similar relationship emerges between the performances of the models based on ground truth and predicted PHOCs. However, compared to the attention approach, the PHOC prediction errors have a stronger impact on the performances of the BIDAF models. In comparison to the line-based approach, the word-based BIDAF model seems to be quite sensitive to erroneous PHOC predictions. Presumably, the line-based model is less sensitive to the recognition errors due to the aggregation on line level. It should be noted, that the performances of our approaches show potential for improvement compared to NLP models working on textual annotations. This gap is likely due to the successful application of transfer learning in the textual domain.

\vspace{.3cm}
\subsubsection{End-to-End Question Answering}
\label{sssec:e2e}
\begin{table}[t]
	\centering
	\caption{End-to-end answer line snippet extraction. Performance measured in DIS.}
	\label{tab:e2el}
	\begin{tabular}{clcc}
		\toprule
		& QA-Approach & HW-SQuAD	& BenthamQA \\ 
		\midrule
		\parbox[t]{2mm}{\multirow{3}{*}{\rotatebox[origin=c]{90}{Pred.}}}&Mathew et al. (rec-free) \cite{Mathew21} & 15.9	& 17.5	\\ 
		& Mathew et al. (rec-based) \cite{Mathew21} & \textbf{59.3}	& 2.5 \\ 
		& BIDAF-Line & 45.0	&  \textbf{37.5} \\	
		\midrule
		\midrule
		\parbox[t]{2mm}{\multirow{2}{*}{\rotatebox[origin=c]{90}{GT}}} &Mathew et al. (rec-based) \cite{Mathew21} & 74.8 &  74.0\\	
		&BIDAF-Line & 45.3 &  55.0 \\	
		
		\bottomrule
	\end{tabular}
\end{table}
In the previous subsections, we have evaluated the individual components of our system. For answering questions in document collections, a combination of those is required. Table \ref{tab:e2el} shows the results for the combination of our document retriever and the line-based BIDAF model as well as approaches from the literature. For this evaluation, the document retrieval approaches extract the top five documents from the collections.

Our approach can clearly outperform the recognition-free method proposed by Mathew et al. \cite{Mathew21}. The recognition-based system from \cite{Mathew21}, however, outperforms our approach on the HW-SQuAD dataset but clearly fails on BenthamQA. This supports the common research outcomes, whereas the performances of textual NLP models are quite high on datasets with low recognition errors, but decrease considerably when the amount of recognition errors rise \cite{Tuselmann21}.  
The results show that the PHOC prediction errors affect the performance of our model, however, it shows a considerably improved robustness compared to recognition-based models.

\section{Conclusions}
In this paper, we present a recognition-free question answering system for handwritten document image collections. The system consists of an attention-based document retriever as well as a question answering approach. Our document retrieval model achieves new state-of-the-art scores for the retrieval task on all considered datasets. For question answering, textual approaches benefit from transfer learning methods and outperform recognition-free approaches on the HW-SQuAD dataset with low word error rates. 
Considering the desired application to historical datasets with presumably no annotated training material, error rates are usually significantly higher. 
As seen on Bentham QA, this leads to a considerable decrease of the QA performance for recognition-based models.
Our experiments show the robustness of our proposed combination of recognition-free retrieval and QA system and that it is able to outperform recognition-free as well as recognition-based methods from the literature. 

\bibliographystyle{IEEEtran}
\bibliography{main}

\end{document}